\documentclass{article}
\usepackage{ijcai16}

\usepackage{times}
\usepackage{epsfig}
\usepackage{graphicx}
\usepackage{amsmath}
\usepackage{amssymb}
\usepackage{mathsymb}
\usepackage{textcomp}
\usepackage{verbatim}
\usepackage{subfigure}
\usepackage{url}
\usepackage{graphicx}
\usepackage{multirow}
\usepackage[super]{nth}
\usepackage{balance}
\usepackage{colortbl}
\usepackage{pgfplots}
\usepackage{pbox}
\usetikzlibrary{patterns}
\pgfplotsset{compat=newest}
\graphicspath{{./}{./Fig/}{./Figs/}{./Fig/eps/}{./Fig/pdf/}}

\usepackage{algorithm2e}

\let\savedalgorithm\algorithm
\let\savedendalgorithm\endalgorithm

\usepackage{algorithm}

\newcommand{\etal}{\textit{et al.\ }}
\newcommand{\eg}{\emph{e.g.,\ }}
\newcommand{\ie}{\emph{i.e.,\ }}
\newcommand{\illshapes}{low quality shapes }
\newcommand{\illshape}{low quality shape }
\newcommand{\shapeill}{shape with low quality }

\title{Effective Multi-Query Expansions: Collaborative Deep Networks for Robust Landmark Retrieval
}
\author{Yang Wang, Xuemin Lin, Lin Wu, Wenjie Zhang}

\begin{document}

\maketitle

\begin{abstract}
Given a query photo issued by a user (q-user),  the landmark retrieval is to return a set of photos with their landmarks similar to those of the query, while the existing studies on the landmark retrieval focus on exploiting geometries of landmarks for similarity matches between candidate photos and a query photo. We observe that the same landmarks provided by different users over social media community may convey different geometry information depending on the viewpoints and/or angles, and may subsequently yield very different results. In fact, dealing with the
landmarks with \illshapes caused by the photography of q-users is often nontrivial and has seldom  been studied. In this paper we propose a novel framework, namely multi-query expansions,
to retrieve semantically robust landmarks by two steps.
Firstly, we identify the top-$k$ photos regarding the latent topics of a query landmark to construct multi-query set so as to remedy its possible \illshape. For this purpose, we significantly extend the techniques of Latent Dirichlet Allocation. Then, motivated by the typical \emph{collaborative filtering} methods, we propose to learn a \emph{collaborative} deep networks based semantically, nonlinear and high-level features over the latent factor for landmark photo as the training set, which is formed by matrix factorization over \emph{collaborative} user-photo matrix regarding the multi-query set. The learned deep network is further applied to generate the features for all the other photos, meanwhile resulting into a compact multi-query set within such space.
Then, the final ranking scores are calculated over the high-level feature space between the multi-query set and all other photos, which are ranked to serve as the final ranking list of landmark retrieval.
Extensive experiments are conducted on real-world social media data with both landmark photos together with their user information to show the superior performance over the existing methods, especially our recently proposed multi-query based mid-level pattern representation method \cite{YXLZ-MM15}.
\end{abstract}


\section{Introduction}\label{sec:introduction}
The popularity of personal digital photography has led to an exponential growth of photos with landmarks. The current social media data set characterize both the landmark photos and associated uploaded user information.
\begin{figure}[htb]
\centering
\includegraphics[width=7cm]{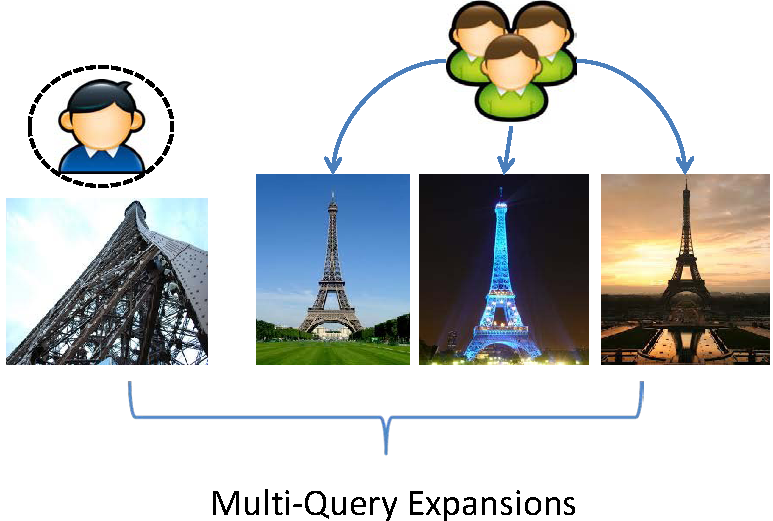}
\caption{\small A q-user has issued a biased landmark photo of Eiffel Tower in Paris in the left most photo. Multiple users with the same latent topic as the q-user are selected to recommend three more landmark photos taken at the same place,
that complement the given query landmark to construct a multi-query set.}\label{fig:example}
\end{figure}
\eg panoramio.com and Picasa Web Album\footnote{picasa.google.com}.
This highly demands for the research in the area of efficient and effective retrieval of photos based on landmarks (\eg tower and churches), namely \emph{landmark retrieval}. Given a query photo, the landmark retrieval returns the set of photos with their landmarks highly similar to that of the query photo.

Unlike the conventional photo retrieval that performs within the low-level feature spaces (\eg color and texture), the landmark retrieval is conducted based on geometry information of landmarks. A number of paradigms \cite{Deorsch:paris,Fang-MM13,im2gps} have been proposed to perform the landmark retrieval under the heterogeneous feature spaces, including the paradigms based on patch level region features \cite{Deorsch:paris}, mid-level attributes \cite{Fang-MM13}, and the combination of low-level features \cite{im2gps}.

Among the existing techniques, there is one critical assumption: a high quality of query photo is always provided; that is, the landmark captured from a query photo always provides a shape with high quality. Nevertheless, such an assumption is not always true in practice.
Indeed, the landmark of a query photo may provide \shapeill due to various reasons, such as personal preference, photography, etc. For example, as illustrated in Fig.~\ref{fig:example}, the left most photo taken by a q-user gives the landmark Eiffel Tower with a very \illshape.
\begin{figure*}[htb]
\begin{center}
\includegraphics[width=12cm]{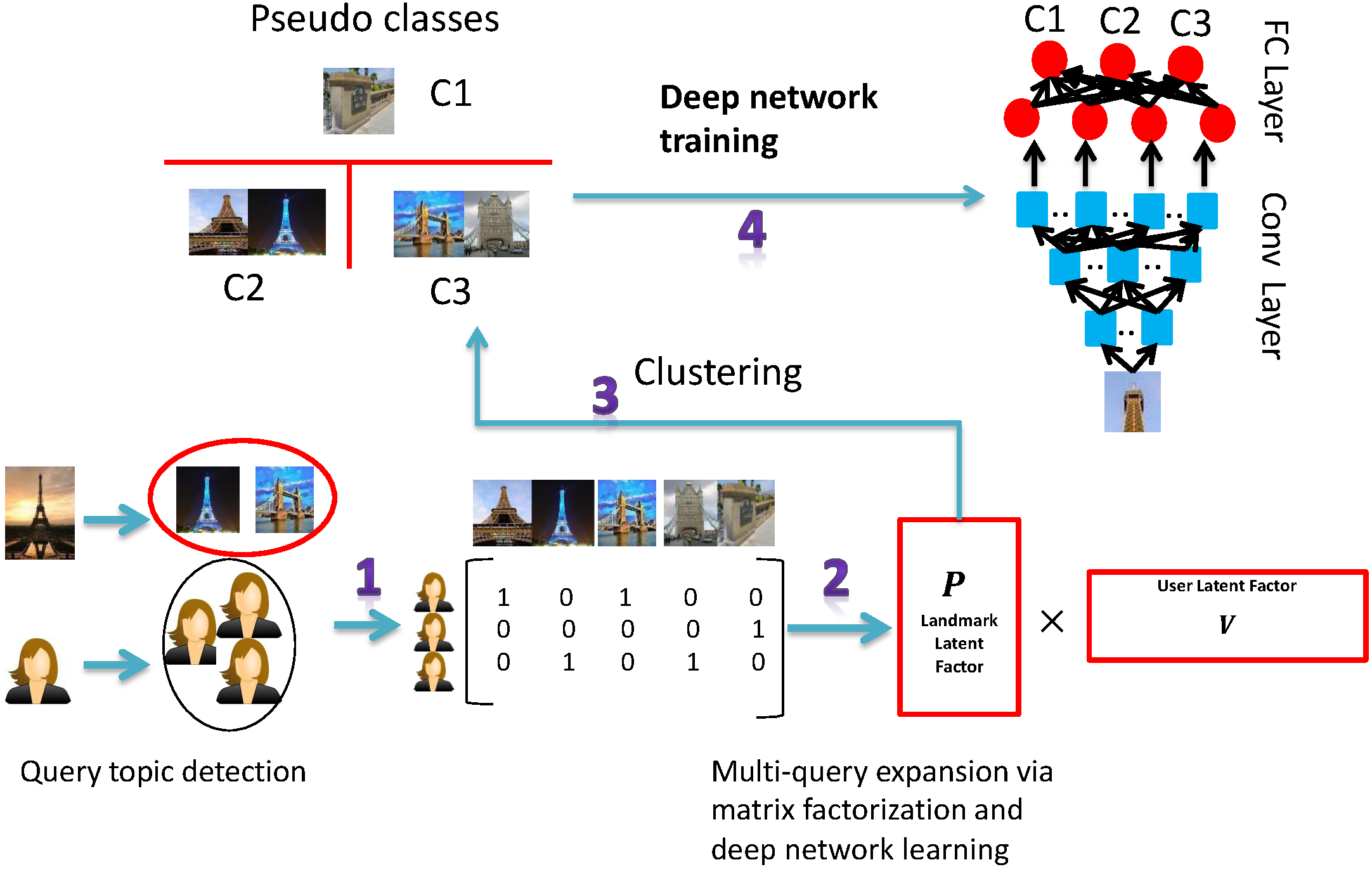}
\caption{\small The flow chart of our framework. Our framework comprises the following main phases: 1) query landmark based topic and user community discovery; 2) select multiple landmark photos from user-query photo matrix to form a multi-query set; and 3) Learning a collaborative deep network based feature representation to characterize the multi-query set and other landmark photos in the database (see Fig.~\ref{fig:deep-arch}), and calculate the ranking score between multi-query set and each photo in the database, leading to the final ranking based retrieval result.}\label{fig:flow-chart}
\end{center}
\end{figure*}
Consequently, the existing methods may not be able to return the set of photos that contain the Eiffel Tower since the geometry quality of the landmark in the query photo is too low.

Motivated by this, in this paper we propose a novel method, depicted in Fig.~\ref{fig:flow-chart}, for a robust landmark retrieval through a novel paradigm based on multi-query expansions over the social media data set with the information for both users and their uploaded landmark photos. Firstly, we propose to identify a set of photos that share the similar latent topics with the query landmark by adopting the Latent Dirichlet Allocation (LDA)  techniques \cite{LDA}.
Then, we propose a localized matrix factorization technique over the user-photo collaborative matrix that encodes the information from the selected photos, where the latent factor regarding landmark photos can be obtained. The k-mean clustering algorithm is subsequently applied to the landmark photo latent factor to generate the clustering result, with each cluster as one pseudo class. \\
\textbf{Remark.} we remark that clustering is performed over latent factors for landmark photo rather than original data objects, as indicated by collaborative filtering methods, such as \cite{YMA:ACMCS}, the latent factors can encode rich underlying similarity so that the ideal clustering result can be achieved.

The deep convolutional Neural Network is then trained under all pseudo classes so that the high-level semantic feature can be extracted via deep network over both query photo and all other landmark photos.

Note that we can select the top-$s$ photos from the multi-query set by calculating the similarity score between the query landmark photo and all other landmark photos within the high-level deep feature space, resulting into a more compact yet encoding similar latent topics as the query landmark photo. We simply combine all the high-level feature representations for all the photos in the multi-query set to be the final representation, which is utilized to calculate the final ranking similarity score over all landmark photos in the database.\\
\textbf{Remark.} We remark that our work is a non-trivial extension of our previous published work \cite{YXLZ-MM15},

\begin{itemize}
\item where a mid-level pattern representation over the expanded multi-query set is proposed for robust landmark retrieval. Following \cite{YXLZ-MM15}, our method exploits not only the information regarding the landmark photos but also the associated user information for multi-query expansion over robust landmark retrieval. Different from \cite{YXLZ-MM15} by learning a mid-level pattern representation for landmark retrieval, we propose to learn high-level semantic feature over the pseudo classes by clustering the latent factors regarding landmark photos yielded from the matrix factorization over the collaborative user-photo matrix.
\end{itemize}

Based on \cite{YXLZ-MM15}, we features the following novel contributions in this paper.
\begin{itemize}
\item Unlike learning the mid-level pattern representations in \cite{YXLZ-MM15}, we learn the high-level deep features to tackle the problem of landmarks in query photos with possibly \illshapes, over both multi-query expansions and landmark photos. To the best of our knowledge, we are the first to learn the deep features for this problem.
\item Unlike \cite{YXLZ-MM15} by exploiting user information to perform multi-query expansion only, We train the deep network over the pseudo classes corresponding to the clusters over the latent factors from the user-photo matrix via matrix factorization.
\item As the matrix factorization over user-photo matrix is categorized as collaborative filtering field, hence we propose a \emph{collaborative} deep network feature learning method for both multi-query set and all other landmark photos, which are further utilized to conduct landmark retrieval within such high-level feature space. We conducted more experiments than \cite{YXLZ-MM15} over real-world landmark photo datasets, validating the effectiveness of our approach.
\end{itemize}

The rest of this paper is structured as follows.  We review the related work in Section ~\ref{sec:related}. Then, we describe our proposed technique in Section ~\ref{sec:technique}. We experimentally validate the performance of our approach in Section \ref{sec:exp}, and conclude this paper in Section \ref{sec:con}.
\section{Related Work}\label{sec:related}
In this section, we mainly review the related work on landmark retrieval and related query argumentation technique.

\subsection{Feature Learning for Landmark Retrieval}
The increasing amount of landmark photos has resulted in numerous methods for landmark retrieval \cite{im2gps,Zhu-MM08,Deorsch:paris,Fang-MM13}. Hays \etal \cite{im2gps} presented a feature matching approach to return the K nearest neighbors with respect to the query landmark photo where a query photo and photos in database are represented by aggregating a set of low-level features to perform landmark retrieval. Zhu \etal \cite{Zhu-MM08} proposed to learn the landmark feature by combining low-level features while assisted with Support Vector Machine (SVM).
In \cite{Deorsch:paris}, a region based recognition method is proposed to detect discriminative landmark regions at patch level, which is seen as the feature for landmark retrieval.  To augment semantic interpretation on landmark representation, Fang \etal \cite{Fang-MM13} presented an effective approach, namely GIANT, to discover both discriminative and representative mid-level attributes for landmark retrieval. Zhu \etal \cite{Landmark-hypergraph} propose a hypergraph \cite{YXLQ-PAKDD14} model for landmark retrieval.

However, these approaches are still using a single query photo for landmark retrieval whilst our approach is focusing on mining robust patterns of landmark photos from an expanded multi-query set. Wang \etal \cite{YXLZ-MM15} proposed to learn a mid-level pattern representations for landmark retrieval over multi-query set, which applied the \textsf{KRIMP} \cite{KRIMP} algorithm and Minimum-description-length (MDL) principle \cite{Grunwald:MDL} to mine the compact pattern representation. Unlike that, we propose a deep network based high-level features for landmark retrieval. As a related problem, several scene categorization methods \eg \cite{PlacesNIPS14,DeepCNN-Scene} via deep feature learning problem is also proposed. Besides the different problem from ours, we also exploited the user information for deep feature learning for landmark retrieval.

There are abundant related work towards feature learning via dictionary learning \cite{TCYB16}. For instance,  Zhu \etal \cite{LingIJCV14} proposed a weakly-supervised cross-domain dictionary learning method is proposed to learn a reconstructive, discriminative and domain adaptive dictionary. To this end, the weakly labeled data is utilized from the source data set to span the coverage of intra-class diversity over the training data to gain discriminative power. However, they have not studied the problem of landmark retrieval.

\subsection{Query Argumentation Technique}
 In \cite{AQE}, a query expansion technique is brought into the visual domain in which a strong spatial constraint between a query image and each result allows for an accurate verification of each return, improving image retrieval performance. This simple method of well-performing query expansion is referred to as Average Query Expansion \cite{AQE} (AQE) where given a query image, images
are ranked using tf-idf scores and corresponding visual words in these images
are averaged together with the query visual words, and this resulting query expanded visual word vector is recast to be a new query to re-query the database.
Observing that AQE is lack of discrimination, Arandjelovic \emph{et al}. \cite{Three-things} enhanced it using a linear SVM to discriminatively learn a weight vector for re-querying yields a significant improvement over the standard average query expansion method, called DQE.
The most related work to ours is \cite{Fernando:iccv13} (PQE), where a query expansion approach for a particular object retrieval is presented. Our method is different from them in terms of multi-query construction. Zhu \etal \cite{LZ:TMM15} propose to perform landmark classification with a hierarchical multi-modal exemplar features. There are also research \cite{Kennedy:www} aiming at developing the feature representations for diverse landmark search.

These methods are commonly based on the idea that those particular multiple queries are manually selected or simply retrieved from top-k similar items whilst we automatically determine helpful queries by exploring the latent topics of query landmark as well as the informative user communities.
Besides, previous query expansion pipelines are not applicable in the context of social media networks, which cannot be addressed by simple variations of methods in literature. To address the limitation, Wang \etal \cite{YXLZ-MM15} proposed a multi-query expansion technique to learn a robust mid-level pattern representation. Based on that, we propose to learn high-level deep feature over multi-query set, which is superior to the mid-level representation in \cite{YXLZ-MM15}.

\subsection{Geo-tagging by Exploring User Community}
Geo-tagging refers to adding geographical identification metadata into various multimedia data such as images \cite{GeoFlickr} in websites, blogs, and photo-sharing web-services \cite{web-a-where,Luo:Geotag,Zheng:www09}. Associating time and location information (latitude/longitude) with pictures can facilitate geotagging-enabled information services in terms of finding location-based news, photos, or other resources \cite{Arslan:mulsys10,Lincvpr10,SeqGeo:iccv09,YXLZQ-MM14}. There are also a number of approaches \cite{Ji:IJCV12} trying to learn the location based visual codebook for landmark search.
Similar to our method, such research area also explore the user communities.
However, this kind of research is apparently different from landmark retrieval studied in this paper.

\subsection{Our Method}
To the best of our knowledge, we are the first to learn the deep network based high-level features over latent factors regarding landmark photos for landmark retrieval. To achieve that, the user information are also explored, which can be easily obtained from the social media data.

Unlike the previous work, inspired by \cite{NIPS12}, our proposed method aims at learning a deep convolutional neural network based high-level features over landmark latent factor obtained by performing matrix factorization over \textbf{\emph{collaborative}} user-photo matrix corresponding to the multi-query set. Then the deep features for both the multi-query set and landmark photos can be generated for further landmark retrieval within high-level deep feature space.

It is well known that learning deep network features plays a crucial role for a wide range of applications. Shao \etal \cite{TNNLS141} proposed a multi-spectral neural networks (MSNN) is by projecting multiple penultimate layer of the deep network, namely multi-columns deep network, into multi-spectral smooth embedding to achieve the complementary multi-view spectral relationship. To this end, an effective multi-view alignment strategy is proposed. As a
by-product, the smooth low-dimensional manifold embedding enables it to be robust to
possible noise corruptions in raw feature representations. A multi-objective genetic program with four-layer structure is proposed in \cite{TNNLS142} to automatically generate domain adaptive global feature descriptors for image classification. A typical yet simple deep network \cite{pcanet} is learned for image classification, by employing PCA to
learn multi-stage filter banks, followed by simple binary hashing and block histograms for
indexing and pooling. Some multi-view feature fusion methods \cite{TIPYang,Yang-cikm13,TNNLS17,Yang-sigir15,Yangijcai16,LY-MM13,LY:IVC17,YangKAIS16} have also been proposed.

Our method features the following non-trivial differences from the above deep network research.
by exploiting the user information associated with the landmark photos to form the landmark photo clusters \ie pseudo-classes, upon which we further learn the deep features for landmark retrieval over such latent factors. We seek the latent cluster representations for user-landmark matrix via matrix factorization, which characterize more useful information than original user-landmark matrix.
Based on the above, given a query landmark photo to be issued, we expand the multi-landmark photo to augment the single query candidate to address the possibly biased query via the Latent Dirichlet Allocation (LDA) for final retrieval.
\section{The Proposed Technique}\label{sec:technique}
In this section, we formally present our technique consists of the following components: The first stage is about latent topic discovery on landmark photos via LDA model, and multi-query expansion based on topic related user groups. The second stage is to train a deep network suitable for landmark photos with user priors such that given input of a multi-query set offered by the first stage, deep features can be extracted regarding both the landmark database as well as the multiple complementary query photos, upon which the deep feature of each photo in the multi-query set are aggregated into overall discriminative representations for retrieval over landmark database.

\subsection{Latent Topics and User Group Discovery}\label{ssec:group-dis}

As stated above, our first stage is to detect the latent topics over landmark photos including query landmark. We first introduce some preliminaries on notations and definitions in LDA model.

\subsubsection{Preliminaries about LDA}

Formally, given a set of users $U$, each of which is associated with their landmark photos. We assume these observed photos can be grouped into clusters as per their latent topics, where each latent topic is related to a set of landmarks; \eg an photo containing the landmark ``Eiffel tower" may belong to the topic ``architecture"; the landmark ``Himalayas" may belong to the topic ``mountain". Suppose a set of topics for a query are detected, each photo can then be modeled as a probabilistic distribution over these topics.
Based on that, it can be seen as a generative model where each user album, a.k.a document $i$, is composed of a number of topics, and the generation for each photo is probabilistically determined by the topics of the album.

It remains a challenge to recommend landmark photos that can complement a query landmark because it is unable to model their similarities by referring to visual appearance which are displaying dramatic changes. To this end, we propose to discover latent topics over the dataset of landmark photos, where the photos sharing the same topics can be clustered together to constitute an augmented multi-query set. We utilized color SIFT descriptors \cite{eccv406} to encode landmark photos, where LDA is further latent topic detection.
In this step, we use Latent Dirichlet Allocation (LDA) \cite{LDA} which allows sets of observations (landmark photos) to be explained by unobserved groups that explain how much of the landmark photos are similar. In our case, photos are regarded as observations that are collected into user albums (a.k.a documents in LDA), and it posits that each document is a mixture of a small number of topics such that each photo's creation is attributable to one of the document's topics. Thus, the topic model of LDA can be used to explain the relatedness of landmark photos which cannot be described reliably by feature descriptors. Once latent topics are discovered w.r.t user album, we can construct a user group that consists of users who upload landmark photos sharing similar topics to the query landmark, from which multi-query expansion can be performed (Section ~\ref{ssec:select}).

\begin{figure}
\begin{center}
\includegraphics[width=7cm]{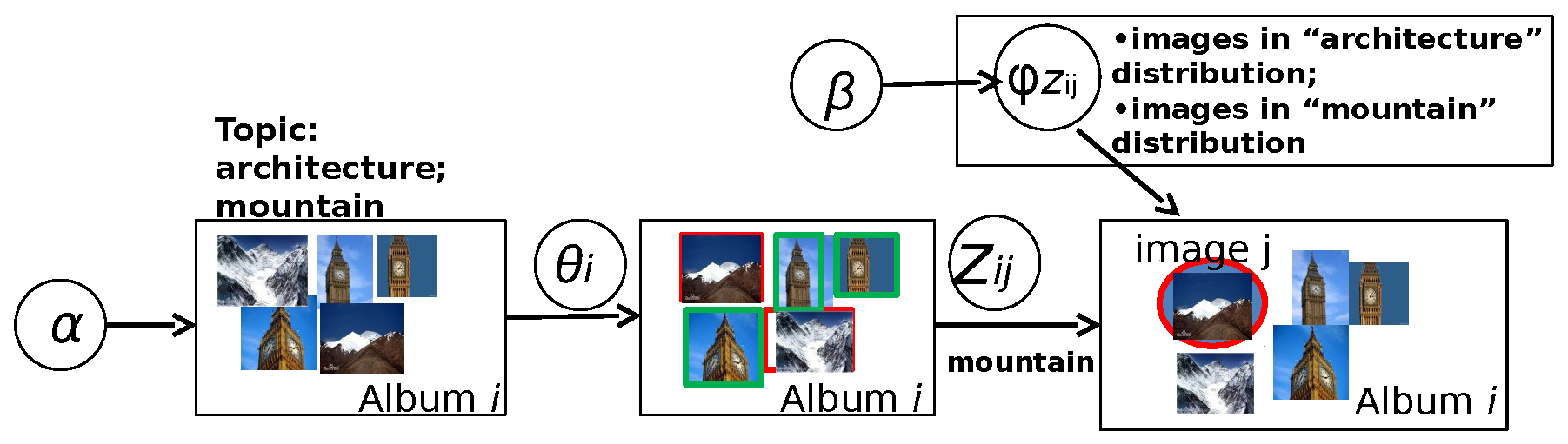}
\caption{ The LDA model of graphical representation on landmark photo distributions. A topic distribution $\theta_i$ (\eg ``architecture" and ``mountain") on album $i$ is chosen by Dirichlet prior $Dir(\alpha)$, and photo distribution $\phi_z$ on topic $z$ is chosen by $Dir(\beta)$. Then each photo $\omega$ from $i$ can be modeled as a distribution on its topic $z_{ij}$ ($\omega \sim Multinomial (\phi_{z_{ij}})$) with topic $z_{ij}$ characterized by $z_{ij} \sim Multinomial (\theta_i)$.}\label{fig:LDA-model}
\end{center}
\end{figure}

Now, we are ready to reformulate the problem of query landmark topic discovery into a LDA model. Let $\alpha$ and $\beta$ be the parameter of the Dirichlet prior on the per-document (album) topic distributions and the per-topic word (photo) distribution. $\theta_i$ is the topic distribution for album $i$, and $\phi_z$ denotes the photo distribution for topic $z$. $z_{ij}$ is the topic for the $j$-th photo in album $i$, and $\omega_{ij}$ is the specific photo. The $\omega_{ij}$ are the only observable variables, and the other variables are latent variables. In fact, $\phi$ is a $Z\times V$ Markov matrix ($Z$ is number of topics considered in LDA, and $V$ is the dimension of the vocabulary, \ie number of photos), and each row of which denotes the photo distribution of a topic. Thus, LDA assumes a generative process for a corpus $D$ consisting of $|D|$ user albums each of length $N_i$:

\begin{enumerate}
\item Choose the number of topics: $Z$
\item Choose $\theta_i \sim Dir(\alpha)$, where $i\in\{1,\ldots,|D|\}$, and $Dir(\alpha)$ is the Dirichlet distribution for parameter $\alpha$
\item Choose $\phi_z \sim Dir(\beta)$, where $z\in\{1,\ldots,Z\}$
\item For each of the photo position $i,j$ where $j\in\{1,\ldots,N_i\}$, and $i\in\{1,\ldots,|D|\}$
\begin{itemize}
\item Choose a topic $z_{ij} \sim Multinomial (\theta_i)$
\item Choose an photo $\omega_{ij} \sim Multinomial (\phi_{z_{i j}})$.
\end{itemize}
\end{enumerate}
Fig.\ref{fig:LDA-model} depicts a graphical model for this representation.

As a result, the output of LDA over $D$, denoted as LDA($D$), can be seen as a partition of $D$ into multiple groups, each of which contains photos characterizing the same latent topics.\\
\textbf{Remark.} To detect the latent topic of the query landmark, we need to perform LDA over the entire photo database, and such process is performed via an off-line fashion in our implementation.

\subsubsection{Detecting Latent Topics for A Landmark Query}\label{sssec:topicsele}

We propose to detect potential latent topics for a landmark query such that highly relevant photos as per each topic w.r..t the query can be clustered, from which a multi-query set can be constructed. Recall that each topic $z$ has the probability of generating photos, and for the query $q$, its probability distribution over $z$ is denoted as $P(q|z)$. Commonly, we set a probability threshold $\lambda$ to decide the candidate topics for the query $q$. That is, $P(q|z) \geq \lambda$ indicates that $z$ could be one candidate topic for $q$. Then, we have two cases regarding the probability threshold $\lambda$:
\begin{enumerate}
\item if $\lambda$ is small, we may generate a lot of candidate topics, while many of them might be non-relevant to $q$.
\item if $\lambda$ is large, the size of topic set would be quite small, which may not be inclusive to involve true topics for $q$.
\end{enumerate}

Thus, we learn a trade-off value for $\lambda$ from empirical studies, which is critical to decide the topic set regarding the query $q$. Once the query topic set is determined, it is plausible to recommend a number of photos sharing the same topics to $q$. To this end, we propose to utilize user communities to assist multi-query set construction. Specifically,  an user is seen to be related to the query user, if he/she uploads photos sharing at least one of the topics associated with an photo uploaded by the query user. Hence, an user-photo matrix can be constructed as $M \in \mathbb{R}^{|U|\times |J|}$ where $U$ and $J$ denote the user and photo set, respectively. $|U|$ and $|J|$ denote the number of users and photos, and $u_q\in U$ represents the query user. In $M$, we have $M(u,j)=1$ if a user $u \in U$ uploaded a photo $j$ into $J$, and $M(u,j) = 0$, otherwise. In what follows, we will elaborate the process of selecting top-$K$ photos to complement the query $q$ to be a robust multi-query set.

\subsection{Multi-Query Expansion}\label{ssec:select}
To discover an appropriate set of photos for query landmark expansion, we perform non-negative matrix factorization \cite{Koren:MF} on the user-photo matrix $M$, where the users (include $u_q$) and relevant photos are detected by LDA described in Section \ref{ssec:group-dis}. By factorizing $M$ into a lower-dimensional latent space, each photo's possibility of being recommended to $u_q$ is computed based on the resulting factorization. Then, photos with $K$ highest relevant scores (high possibility to share same topics w.r.t the query) are selected, together with the original query, to form a robust multi-query set.

Specifically, we perform the non-negative matrix factorization on $M$ by minimizing the objective function below.
\begin{equation}\label{eq:MF}
\min_{P,V} ||M-P V||_F, P \geq 0, V \geq 0,
\end{equation}
where $||\cdot||_F$ denotes the Frobenius norm. Two low-rank factor matrices $P\in R^{|U|\times L}$ and $V\in R^{L\times |J|}$ are obtained by solving Eq.\eqref{eq:MF}, where $P$ models the mapping of users in the low-dimensional latent space with $L$ dimensions, and $V$ defines the mappings of photos to the same latent space. That is, each user $u$ is represented as the $u$-th row vector of $P$ denoted by $P(u_q,\cdot)$, while each photo $j$ is encoded by the $j$-th column vector of $V$ denoted by $V(\cdot,j)$. We reformulate Eq.~\eqref{eq:MF} into the following optimization problem
\begin{equation}\label{eq:MFREF}
\min_{P,V} ||M - P V||_F^2 + \beta(||P||_F^2 + ||V||_F^2),
\end{equation}
where $\beta$ is the balance parameter, and the stochastic gradient descent \cite{SCDE} is applied to solve Eq. \eqref{eq:MFREF}.

As presented in our earlier work \cite{YXLZ-MM15}, relevant photos can be selected by computing their inner product with the query user vector in the factorized space. Specifically, for $u_q \in U$, a confidence score can be computed to determine if photo $j$ will be recommended to $u_q$ through the inner product of $P(u_q,.)$ and $V(.,j)$:
\begin{equation}\label{eq:select}
S(u_q,j)=<P(u_q,\cdot),V(\cdot,j)>.
\end{equation}

Through the computation of Eq. \eqref{eq:select}, we can select $K$ photos with high confidence scores to construct a robust multi-query set $Q$ that describe a landmark of interest from different aspects. To improve the landmark recognition, faithfully representative descriptions for categorical landmark photos are needed. To leverage multiple photos in $Q$ and generate descriptive representation for a landmark, an effective mid-level pattern mining strategy is proposed in our earlier work \cite{YXLZ-MM15} where frequent patches are discovered by principled minimum description length. However, this pattern representation is obtained via an unsupervised fashion, which carries no semantically categorical information in landmark photos.

To learn high-level discriminative representations for landmark photos, we propose to learn deep features for $Q$ through the powerful convolutional neural networks (CNNs) \cite{NIPS12} which have shown impressive performance in a variety of recognition tasks.  In our case, a trainable deep network can be fine-tuned on the landmark data to generate semantically high-level features regarding landmark samples in categories. We describe the design and training of deep network in Section \ref{ssec:deep_network}. Once multiple deep features for landmark photos in $Q$ are learned, we aggregate them into overall feature representation to describe a landmark of interest, and perform similarity search between the query and each landmark photo  in database with their high-level features. In our paper, we use Euclidean distance to calculate similarity scores.\\
\textbf{Remark.} One may think only LDA can make the satisfied retrieval results. However 1) the LDA can only detect the landmarks sharing the similar latent topics above a probability threshold. 2) Even the candidates within the similar topics, the final retrieved landmark candidates still need to be ranked according to their landmark relevance to achieve a satisfied accuracy based on evaluation metric. To this end, we need to learn a strong deep feature representations over both multi-query set and landmark photos for the final retrieval.

\subsection{Deep Landmark Feature Learning on Collaborative Convolutional Neural Networks (C-CNNs)}\label{ssec:deep_network}

\begin{figure*}[hbt]
\centering
\begin{tabular}{c}
\includegraphics[width=13cm,height=2cm]{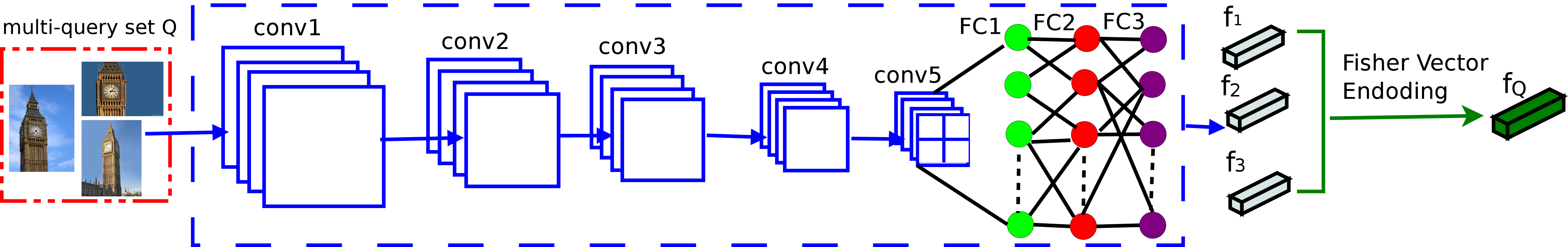}\\
\end{tabular}
\caption{ Given a multi-query set $Q$ in testing, fine-tuned parameter set of CNN-F are applied to feed-forwardly extract high-level discriminative features for each photo in $Q$. The resulting multiple feature vectors are encoded using Fisher vector encoding to generate an overall feature vector which can best describe the landmark of interest in a robust and discriminative way.}\label{fig:deep-arch}
\end{figure*}

The query expansion step produces a set of query photos that are describing the landmark of interest in different aspects. To combine these complementary priors, we focus on constructing photo representations, \ie encoding functions $\phi$ mapping each query photo $I$ to a vector $\phi (I)\in \mathbb{R}^d$, and an effective combination method to integrate multiple vectors into an overall feature vector.
Since these query photos are displaying landmarks with visual variance, we employ the Convolutional Neural Networks (CNN) \cite{NIPS12} with several layers of non-linear feature extractors to generate high-level deep representations.

\subsubsection{Deep Representations with Pre-training}

Our deep representations are inspired by the success of CNN in photo classification\cite{NIPS12}. As shown in \cite{Decaf,UnderstandCNN}, the vector of $\phi(I)$ of a deep CNN, learned on large dataset such as photoNet \cite{ImageNet}, can be used as a powerful photo descriptor applicable to other datasets. Here we adopt a fast architecture (CNN-F \cite{Devil}) which is similar to the one used by Krizhevsky \etal \cite{NIPS12}. It comprises 8 learnable layers, 5 of which are convolutional, and the last 3 are fully-connected. The input photo size is $224\times 224$. Fast processing is ensured by the 4 pixel stride in the first convolutional layer. The main difference between CNN-F and that of \cite{NIPS12} are the reduced number of convolutional layers and the dense connectivity between convolutional layers. The structure of CNN-F is shown in Table \ref{tab:convnet}.

\begin{table*}[t]\tiny
  \centering
  \caption{ The CNN-F architecture. It contains 5 convolutional layers (conv 1-5) and three fully-connected layers (FC 1-3). The details of each convolutional layer are given in three sub-rows: the first specifies the number of convolution filters and their receptive field size as ``num $\times$ size $\times$ size''; the second indicates the convolution stride (``st.'') and spatial padding (``pad''); the third indicates if Local Response Normalisation (LRN)  is applied, and the max-pooling downsampling factor.}  \label{tab:convnet}
  {
  \begin{tabular}{l|c|c|c|c|c|c|c|c}
  \hline
\hline
    Arch.  & conv1  &  conv2 & conv3 & conv4 & conv5 & FC1& FC2 & FC3\\
  \hline\hline
CNN-F (F-Net) & \pbox{5cm}{$64\times 11\times 11$ \\ st. 4, pad 0 \\LRN, $\times$2 pool} & \pbox{5cm}{$256\times 5\times 5$ \\ st. 1, pad 2 \\LRN, $\times$2 pool}  & \pbox{5cm}{$256\times 3\times 3$ \\ st. 1, pad 1 \\- }  & \pbox{5cm}{$256\times 3\times 3$ \\ st. 1, pad 1 \\ -} & \pbox{5cm}{$512\times 3\times 3$ \\ st. 1, pad 1 \\ $\times$2 pool}  & 4096 dropout & 4096 dropout & $C$-way soft-max\\
\hline
  \hline
  \end{tabular}
  }
\end{table*}

\subsubsection{CNN Fine-tuning on the Target Dataset}
It was demonstrated \cite{RichFeature} that fine-tuning a pre-trained CNN on the target data can significantly improve the performance, and we consider this scenario to obtain photo features which are dataset-specific to landmark dataset. In our framework, we fine-tuned CNN-F using the landmark dataset where fine-tuning was carried out using the CNN pre-trained parameters on ILSVRC \cite{ILSVRC}. We employ data augmentation in the form of cropping and flipping. CNN requires photos to be transformed to a fixed size ($224\times 224$), and hence the photo is downsized so that the smallest dimension is equal to 224 pixels. Then $224\times 224$ crops are extracted from the four corners and the centre of the photo. These crops are then flipped about the $y$-axis, producing 10 perturbed samples per input photo.

Note that the last fully-connected layer (FC3) has output dimensionality equal to the number of classes, which differs between datasets. To this end, we cluster the landmark photos within the latent factor with $L$ dimensional space into \emph{C} clusters via \emph{K}-means, which naturally corresponds to the \emph{C} pseudo-classes regarding supervision information (see Fig.~\ref{fig:flow-chart}). Clustering landmark photos  in their factorized latent space instead of using their original category priors suggests user related supervision signals. As the latent factors regarding the landmark photos yielded by matrix factorization follows the principle of collaborative filtering \cite{YMA:ACMCS} in recommendation system, we name the feature learning procedure as deep landmark feature learning on \emph{Collaborative} Convolutional Neural Networks (denoted as C-CNNs). In this way, latent factors characterize rich similarity information among landmark photos, which can facilitate the clustering process to produce accurate clusters as supervision signals. The essence of the proposed C-CNNs is to map latent factors to the landmark photo pixels, such that the user-photo correspondence can be encoded. However, latent factors regarding a landmark query could be noisy. To this end, softmax function is adopted \cite{YTICML13} on the top layer of the deep network to classify the membership of input photo probabilistically.

\subsubsection{Fisher Vector Encoding on Multi-Query Set}
Once we have obtained trainable parameters of C-CNNs, the multi-query set $Q$ can be transformed through the network to produce multiple high-level feature vectors $Q=\{x_1,\ldots,x_{|Q|}\}$. To aggregate these vectors into an overall representation that best describes the landmark of interest, we employ Fisher Vector to encode them with high-order statistics \cite{FisherVectorIJCV} (See Fig. \ref{fig:deep-arch}).  An alternative aggregation strategy is average-pooling, which is formulated as:
$\widehat{x}=\frac{1}{|Q|}\sum_{i=1}^{|Q|} x_i.$
However, average-pooling is unable to capture the correlation of multiple feature vectors in regards to a particular landmark. In fact, landmarks within scene photos could exhibit dramatically different appearances, shapes, and aspect ratios. To distinguish one landmark category from another, it is much desired to harvest discriminative and representative landmark-specific object parts, which are collected in the multi-query set $Q$. To this end, we use Fisher Vector (FV) \cite{FisherVectorIJCV} to aggregates multiple deep feature vectors into a high dimensional non-linear representation that is suitable for classifier. FV can effectively address geometric variance in scene photos, and can be combined with CNNs \cite{DeepFisherKernel,FVsMeetNNs} to improve classification performance. We quantitatively evaluate the combination of C-CNNs with FV or average-pooling, as shown in \ref{ssec:compare_baseline}.

The Fisher vector encoding $\Phi$ of a set of features is based on  fitting a parametric generative model such as the Gaussian Mixture Model (GMM) to the features, and then encoding the derivatives of the log-likelihood of the model with respect to its parameters \cite{GenerativeNIPS1998}. It has been shown to be a state-of-the-art local patch encoding technique \cite{FisherVectorIJCV,DevilBMVC2011}. Our base representation of a query photo is a set of deep feature vectors corresponding to multi-query set $Q$. The descriptors are first PCA-projected to reduce their dimensionality and decorrelate their coefficients to be amenable to the FV description based on diagonal-covariance GMM.
Given a GMM with $G$ Gaussians, parameterized by $\{\pi_g, \mu_g, \sigma_g, g=1\ldots, G\}$, the Fisher vector encoding leads to the representation which captures the average first and second order differences between the features and each of the GMM centres \cite{ImproveFisherKernel,DeepFisherNetworks}. For a descriptor $x\in \mathbb{R}^D$, we define a vector $\Phi(x)=\left[\varphi_1(x),\ldots, \varphi_G(x), \psi_1(x),\ldots,\psi_G(x)\right]\in \mathbb{R}^{2GD}$. The subvectors are defined as
\begin{equation}\label{eq:FV}
\begin{split}
&\varphi_g (x)=\frac{1}{\sqrt{\pi_g}} \gamma_g (x) \left(\frac{x-\mu_g}{\sigma_g}\right)\in \mathbb{R}^D,\\
& \psi_g(x)=\frac{1}{\sqrt{2\pi_g}}\gamma_g(x) \left(\frac{(x-\mu_g)^2}{\sigma_g^2}-1\right) \in \mathbb{R}^D,
\end{split}
\end{equation}
where $\{\pi_g, \mu_g, \sigma_g\}_g$ are the mixture weights, means, and diagonal covariance of the GMM, which are computed on the training set; $\gamma_g(x)$ is the soft assignment weight of the feature $x$ to the $g$-th Gaussian. In this paper, we use a GMM with $G=256$ Gaussians.

To represent a query set $Q$, one averages/sum-pool the vector representations of all descriptors, that is, $\Phi(Q)=\frac{1}{|Q|}\sum_{i=1}^{|Q|} \Phi(x_i)$. The Fisher vector is further processed by performing signed square root and $\ell_2$ normalization on each dimension $d=1,\ldots, 2GD$:
\begin{equation}\label{eq:fisher_norm}
\bar{\Phi}_d(Q)=(sign~~ \Phi_d(Q)) \sqrt{|\Phi_d(Q)|}/ \sqrt{||\Phi(Q)||_{\ell_1}}
\end{equation}
For simplicity, we refer to the resulting vectors from Eq.\eqref{eq:fisher_norm} as Fisher vectors and to their inner product as Fisher kernels.
\section{Experiments}\label{sec:exp}

In this section, we present extensive experimental results by comparison to a variety of baselines and state-of-arts to evaluate the effectiveness of our approach.

\subsection{Datasets}

Two datasets are constructed by collecting landmark photos from \textbf{Flickr} and \textbf{Picasa Web Album}.
They are suitable for landmarks retrieval because they contain both user information and corresponding landmark photos.

\begin{figure}
\centering
\begin{tabular}{c}
\includegraphics[height=3.5cm]{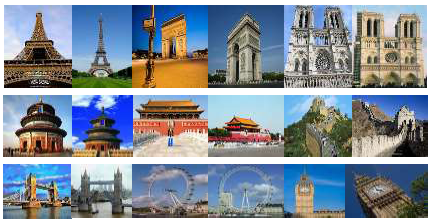}\\
\end{tabular}
\caption{Examples of landmark photos from our collected Flickr dataset.}
\label{fig:examples}
\end{figure}

\begin{itemize}
\item Flickr. We use the Flickr API to retrieve landmark photos taken at a city posted by a large number of users.  We sort out 11 cities: London, Paris, Barcelona, Sydney, Singapore, Beijing, Tokyo, Taipei, Cairo, New York city, and Istanbul. In each city, such as Paris, we obtained photos by querying the associated text tags for famous landmarks such as ``Paris Eiffel Tower" or ``Paris Triomphe" \footnote{In Paris, 12 queries were used to collect photos from Flickr: La Defense Paris, Eiffel Tower Paris, Hotel des Invalides Paris, Louvre Paris, Moulin Rouge Paris, Musee d'Orsay Paris, Notre Dame Paris, Pantheon Paris, Pompidou Paris, Sacre Coeur Paris, Arc de Triomphe Paris, Paris}. Example photos containing a variety of landmarks from this dataset are shown in Fig. \ref{fig:examples}.

\item Picasa Web Album. Providing rich information about interesting tourist attractions, this source contains a vast amount of GPS-tagged photos uploaded by users who have visited the landmarks, along with their text tags.
We manually download a fraction of photos and their user information on 6 cities: London, Paris, Beijing, Sydney, Chicago, and Barcelona.
\end{itemize}

The statistics for user information and their uploaded photos over the two datasets are summarized in Table \ref{tab:sta-dataset}.

\begin{table}
\centering
\small
\caption{Statistics of datasets.}
\begin{tabular}{|c|c|c|} \hline
Dataset & \textbf{Flickr} & \textbf{Picasa Web Album}\\ \hline
\# Landmark & 55 & 16\\ \hline
\# photo per landmark & nearly 1,000 & 100 $\thicksim$ 300\\ \hline
\# Total photo & 49,840 & 4,100\\ \hline
\# User & 7,332 & 577\\
\hline\end{tabular}\label{tab:sta-dataset}
\end{table}

\textbf{Remark.} It is noteworthy to remark the followings
\begin{itemize}
\item For our data set, we don't follow the assumption of selecting the same landmark photo from the different users, as also mentioned by our previous conference paper \cite{YXLZ-MM15}, such multi-query expansion comprising the same redundant landmark photos is not informative. Actually, for our practical implementations, we seldom meet such scenario.
\item It is more commonly a scenario that the same landmark with different capturing conditions such as varied capturing angles. Given a possible biased query landmark photo, we just attempt to choose the other more photos that characterize the same landmark yet with a more ideal conditions to reflect its ground-truth.
\item However, we have to admit that each photo set as we adopt characterizes the same ground-truth landmark with a relative smaller portions of biased landmark photo, which ensure to validate the effectiveness of our multi-query expansion technique.
\end{itemize}

\subsection{Training and validation}

For all datasets, we split them with the training, validation and testing  70$\%$, 10$\%$ and 20$\%$ respectively. The training set is utilized for C-CNN feature learning and the testing evaluates the learned feature representation. We also leave a small portion of training set for validation. As aforementioned, we perform matrix factorization over the entire training data by solving Eq.~\eqref{eq:MFREF} to estimate the parameter $\beta$. Note that we adopt the root-mean-square error \cite{KDD11cup} as the validation metric to estimate the optimal parameter $\beta$ for Eq.~\eqref{eq:MFREF}. The validation process shows that $\beta$=0.05 and $L$=200 can reach satisfied performance in terms of root-mean-square error. Besides, we set the cluster number over latent factors $C$=1000 for C-CNN feature learning.

\subsection{Settings}
\paragraph{Evaluation Metric} We choose precision-recall as the metrics to evaluate both our model and state-of-the-art methods over top-100 retrieval ranked list w.r.t each query landmark. In each landmark, we issue 20 queries and an average precision score is computed for all queries, which are averaged to obtain a mean Average Precision (mAP) for the each landmark category.

\paragraph{Baselines}
We consider several variants of our model to understand each component of the proposed C-CNN. Also, some recent deep scene classification methods are considered as strong baselines:
\begin{itemize}
\item CNN-F+Average: The CNN-F model is fine tuned on original class labels, and then average pooling is used to generate the overall feature vector.
\item CNN-F+FV: The CNN-F model is fine tuned on original class labels, and then Fisher Vector encoding is used to generate the overall feature vector.
\item C-CNN+Average: The proposed collaborative CNN is fine tuned on pseudo labels from user supervision, and then average pooling is used to generate the overall feature vector regarding the multi-query set.
\item Ours (C-CNN+FV): The architecture is the same as C-CNN except that Fisher Vector encoding is employed to produce the overall feature representation.
\item Deep Places Features: This is a strong deep baseline which uses CNN to learn deep features from the largest scene-centric database, namely Places Database \cite{PlacesNIPS14}. The Places database contains over 7 millions labeled pictures of scenes from which the CNN features can achieve state-of-the-art results.
\item MetaObject-CNN \cite{DeepCNN-Scene}: This method is to learn discriminative features for scene classification by fine-tuning CNN on pre-generated patches containing objects and parts that frequently occur in photos for scene category.
\end{itemize}

\paragraph{Competitors} In our experiment, we consider nine competitors listed below. In their implementation of training, SIFT descriptor \cite{Lowe:sift} is used as a local descriptor due to its excellent performance in object recognition \cite{Boureau:mid-level}. Specifically, we adopt a dense sampling strategy to select the interest regions from which SIFT descriptors are extracted.

\begin{itemize}
\item \textsf{K-NN} \cite{im2gps}:  A feature matching approach to return the $K$ nearest neighbors with respect to the query landmark photo where a query photo and photos in database are represented by aggregating a set of low-level features to perform landmark retrieval.

\item \textsf{LF+SVM}: Low-level features \cite{Zhu-MM08} combined with SVM.
\item  \textsf{DRLR} \cite{Deorsch:paris}: A region based location recognition method that detects discriminative regions at the patch-level.
\item \textsf{GIANT} \cite{Fang-MM13}: A method to discover geo-informative attributes that are discriminative for location recognition.
\item \textsf{AQE} \cite{AQE}: Average Query Expansion method that proceeds as follows: given a query region, it ranks a list of photos using tf-idf scores. Bag-of-Word vectors corresponding to these regions are averaged with BoW vectors of the query, resulting in an expanded vector used to re-query the database.

\item \textsf{DQE} \cite{Three-things}: Discriminative Query Expansion that enriches a query in the exactly same way as AQE. It considers photos with lower tf-idf scores as negative data to train a linear SVM for further rankings and retrievals.

\item \textsf{PQE} \cite{Fernando:iccv13}: A Pattern based Query Expansion algorithm that combines top-K retrieved photos with a query to find a set of patterns.

\item \textsf{PAMQE} \cite{YXLZ-MM15}: A query expansion method by yielding a mid-level pattern representation over expanded multi-query set via LDA model against the user-photo matrix. To ensure its generalization where test data differs from the dataset used to generate the quantization, we use a different dataset, the \textbf{Oxford} dataset\footnote{http://www.robots.ox.ac.uk/vgg/data/oxbuildings/}, to construct the vocabulary book from which local bag-of-words are encoded as representations \footnote{The Oxford Building dataset contains 5,062 photos, which is a standard set of particular objects for retrieval. The reason for choosing Oxford landmarks is that the photos exhibit scenes similar to, rather than identical landmarks, those in the two test databases (e.g., buildings have much similarities in their architectural styles).}.

\item \textsf{MMHG} \cite{Landmark-hypergraph}: This is an effective modeling scheme to characterize the associations between landmark photos by using multiple hypergraphs to describe high-order relationship from different aspects.
\end{itemize}

\subsection{Tuning Parameters}\label{ssec:para-learn}

In experiments, we tune three parameters as follows,
\begin{itemize}
\item K: the number of queries after matrix factorization in the multi-query set.
\item $\lambda$:  the threshold to determine the latent topic set w.r.t the query landmark.
\item $L$: the reduced dimension or number of latent factor in matrix factorization.
\end{itemize}
Following \cite{YXLZ-MM15}, we set K=40, $\lambda$=0.4 and $L$= 64 for a fair comparison. We set $s$=20 to select top 20 photos, which characterize the 20 most similar items to the query landmark described by the C-CNN high-level features, to construct a compact multi-query set for landmark retrieval.

\begin{figure}[t]
\centering
\includegraphics[width=1.5cm]{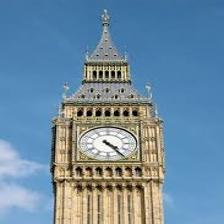}
\includegraphics[width=2cm]{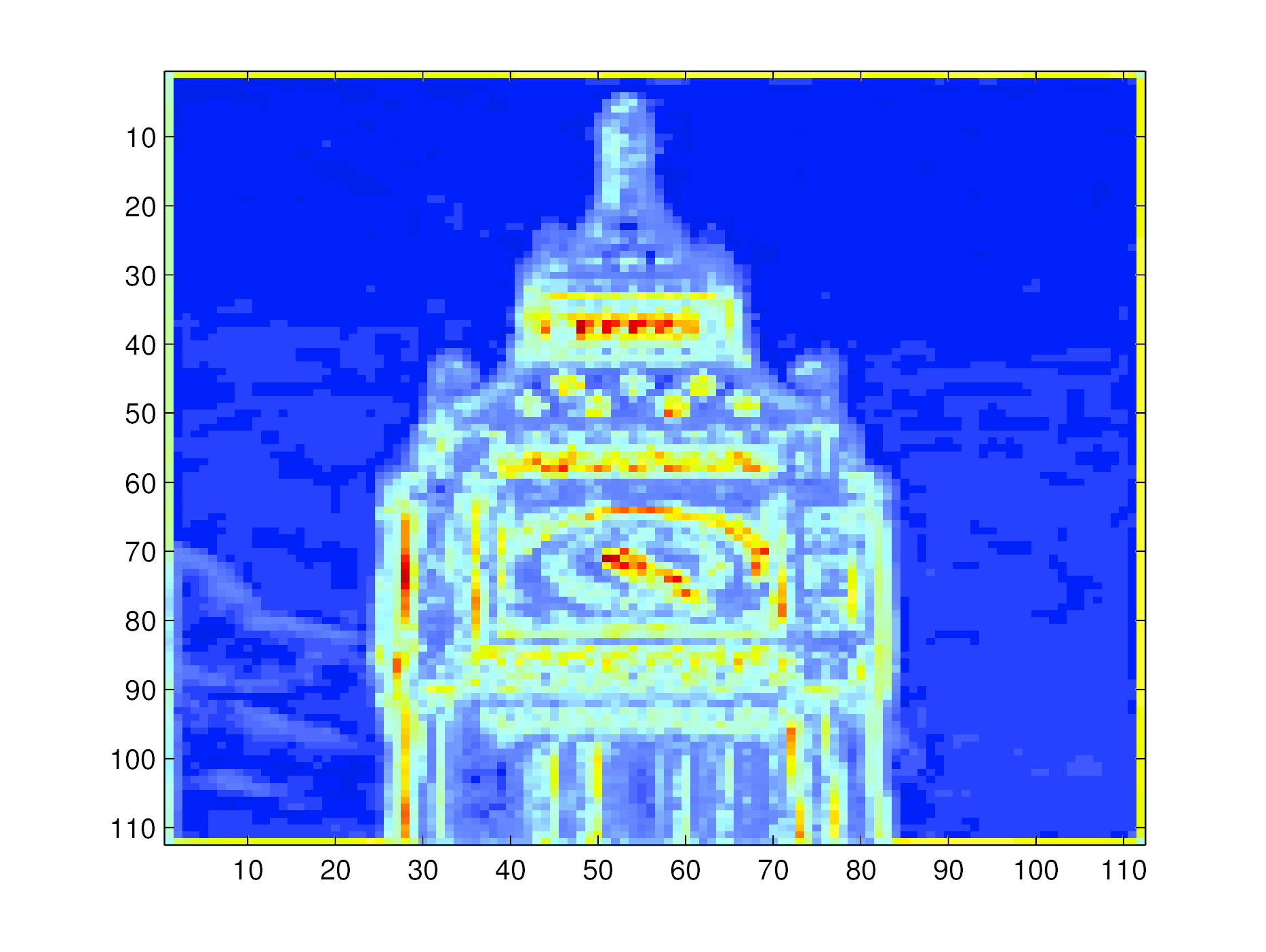}
\includegraphics[width=1.5cm]{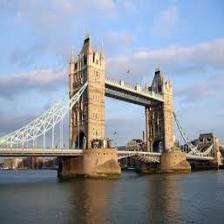}
\includegraphics[width=2cm]{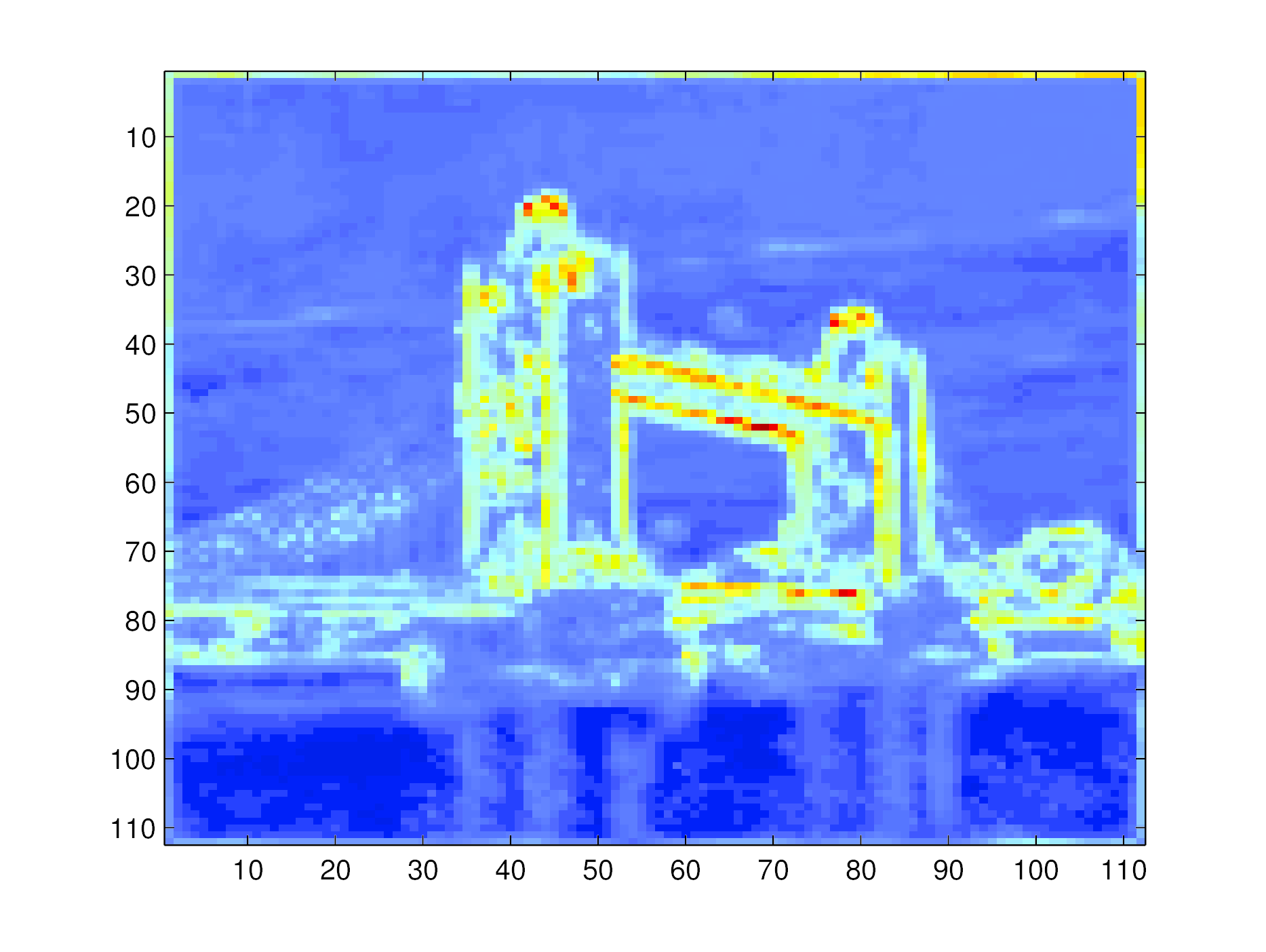}\\
\includegraphics[width=1.5cm]{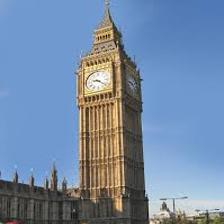}
\includegraphics[width=2cm]{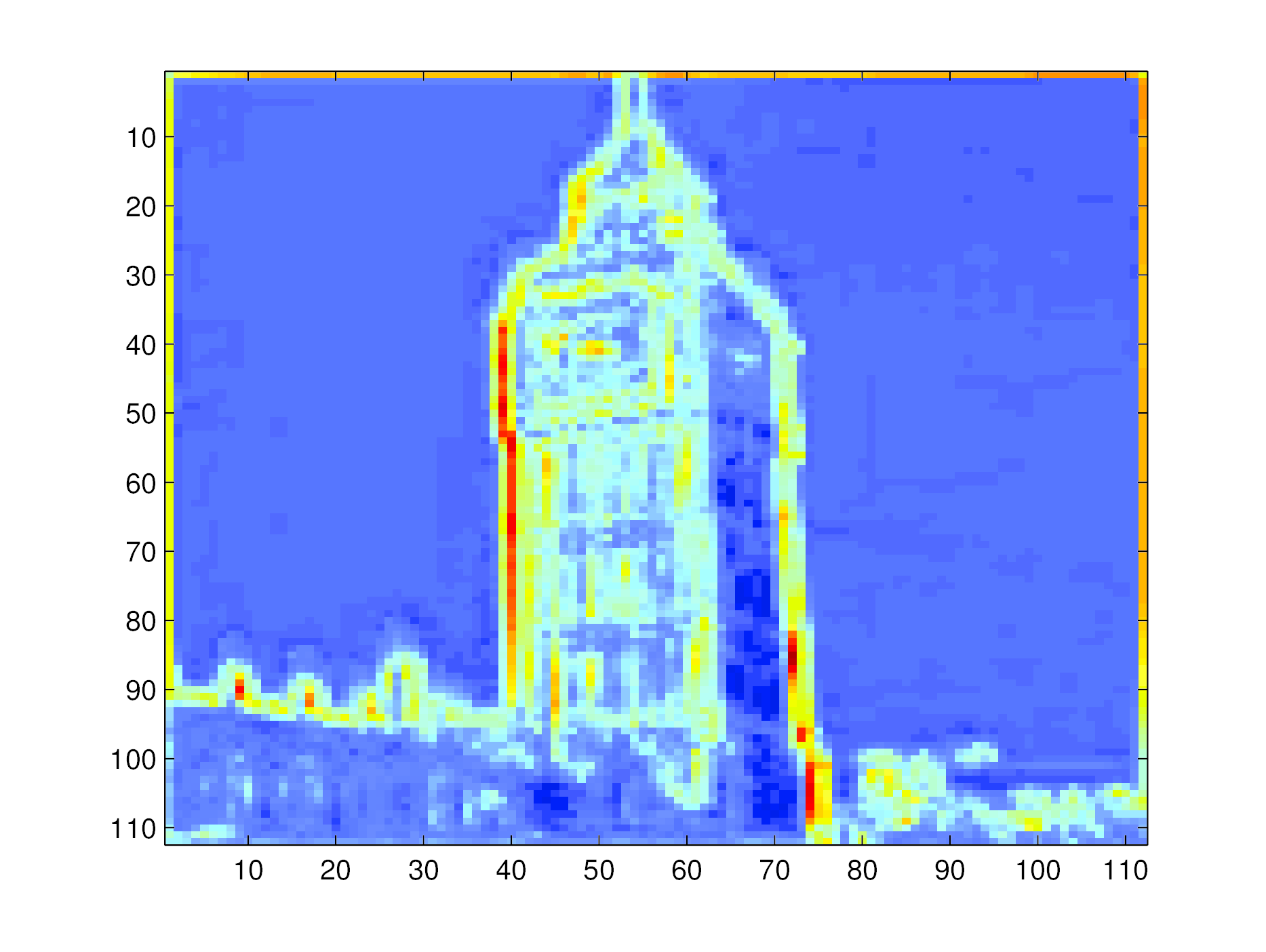}
\includegraphics[width=1.5cm]{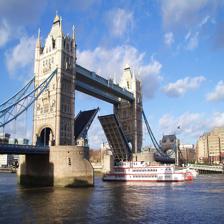}
\includegraphics[width=2cm]{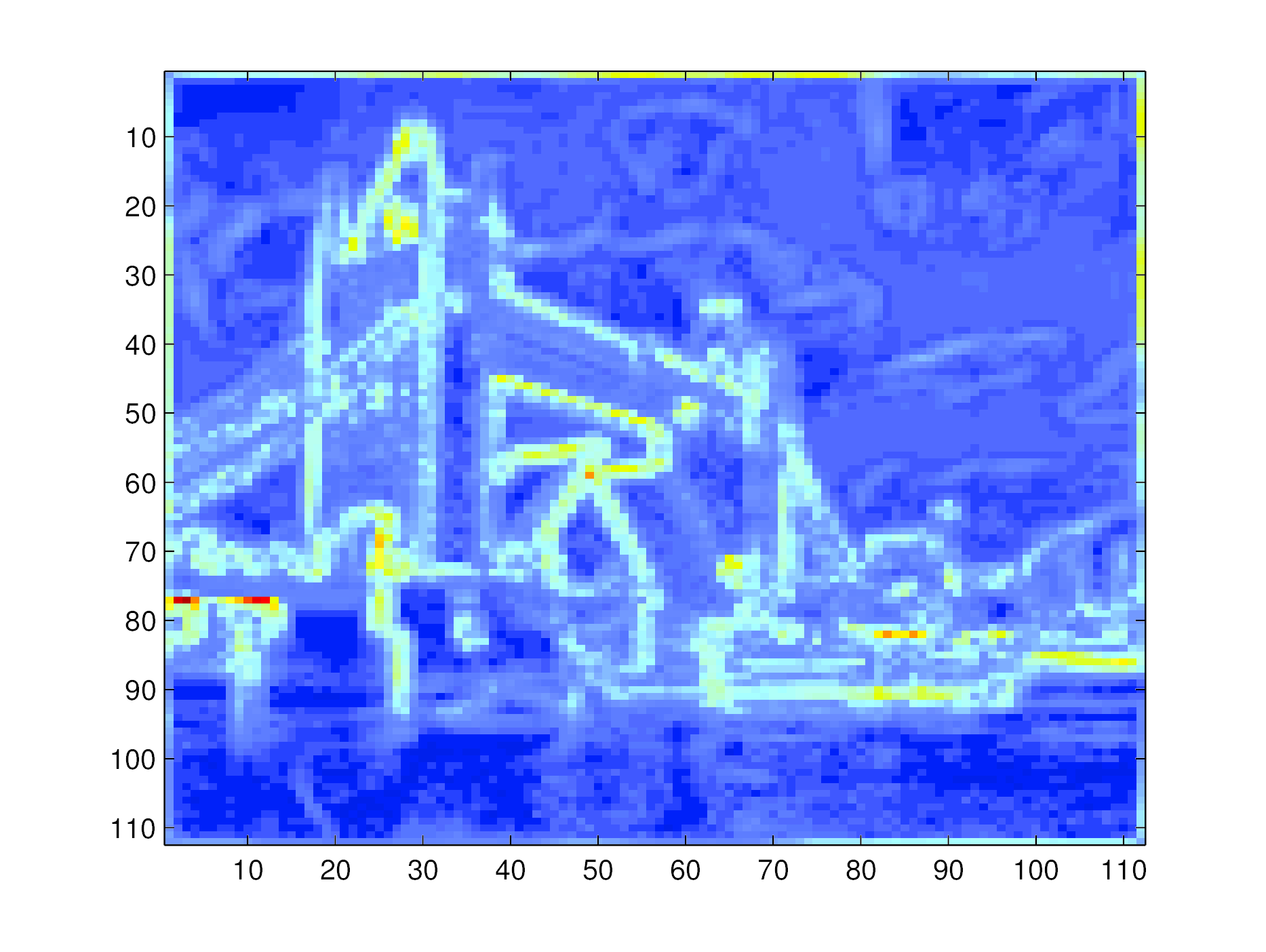}\\
\includegraphics[width=1.5cm]{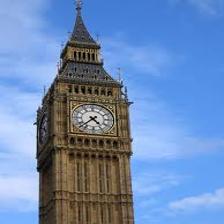}
\includegraphics[width=2cm]{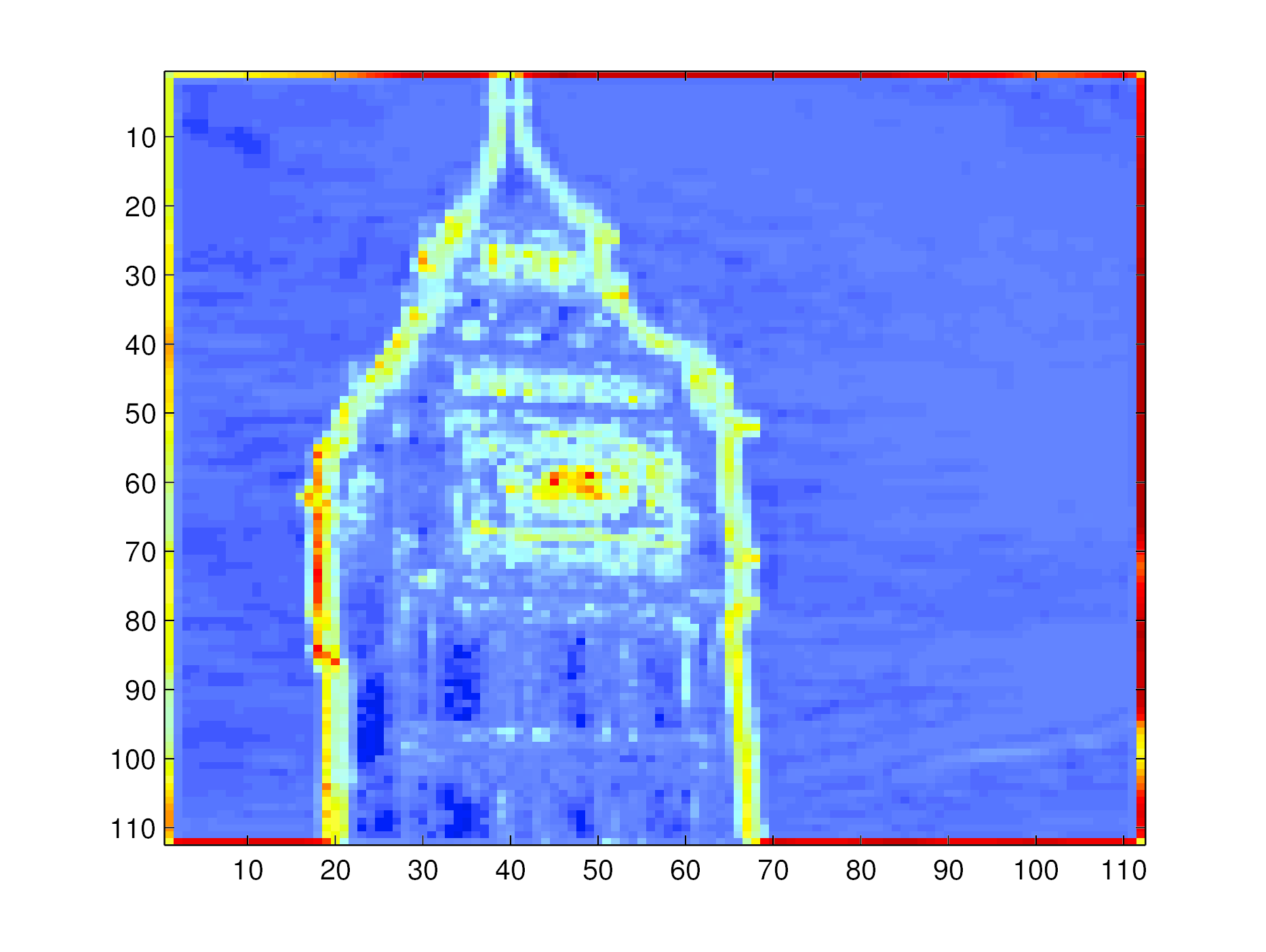}
\includegraphics[width=1.5cm]{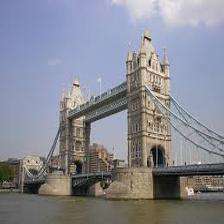}
\includegraphics[width=2cm]{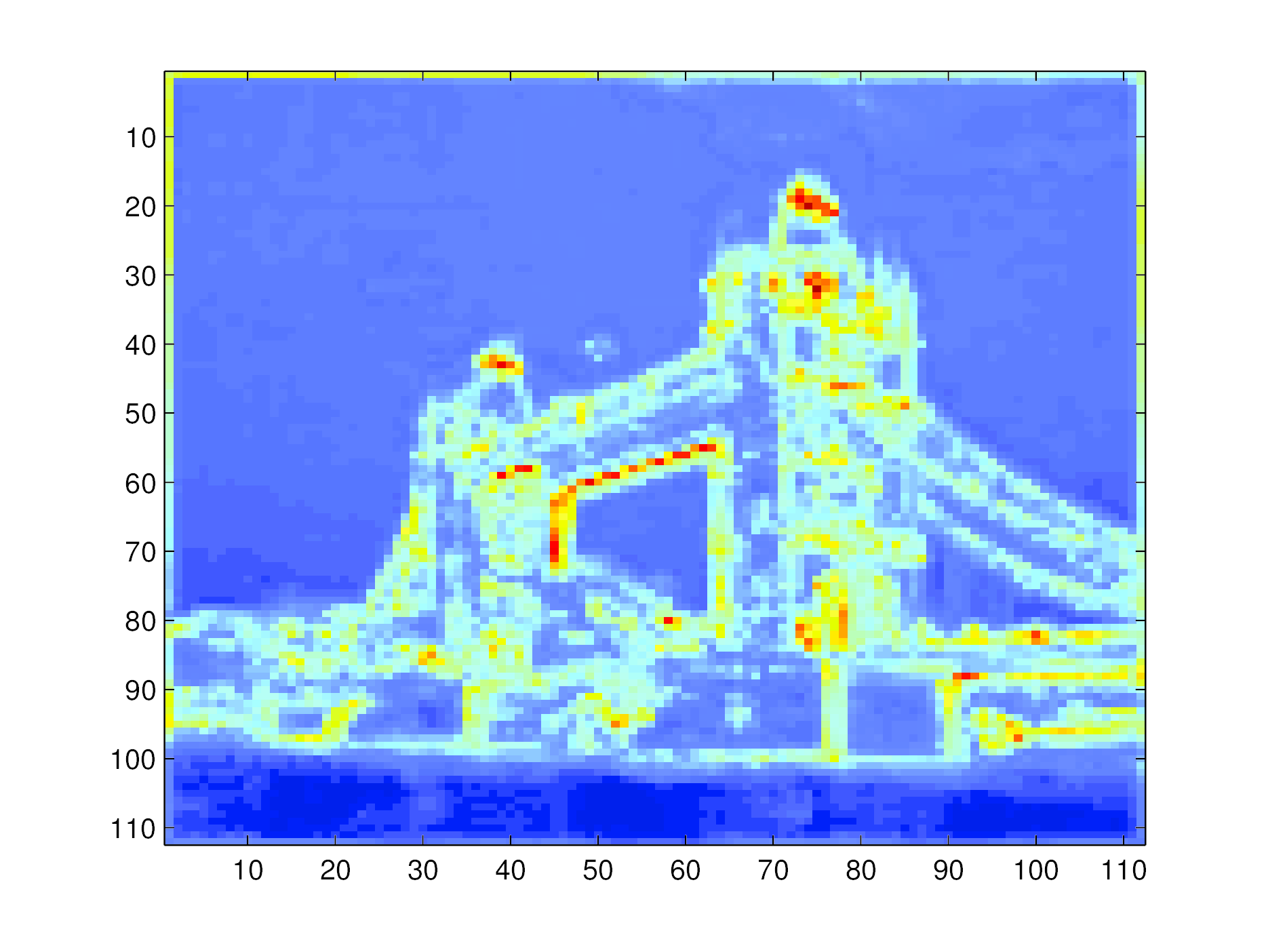}\\
\caption{An example consists of the original test photo (left) and the corresponding heatmap represented the highest response regions during the prediction.}\label{fig:feature_map}
\end{figure}

\begin{table*}
\tiny
\caption{The mAP ($\%$) values of selected landmarks by different approaches against Flickr dataset.}
\begin{tabular}{c|c|c|c|c|c|c|c|c|c|c|c|c}\hline
\small
\multirow{2}{1cm}{Method} & London & Triumphal & Sagrada & Sydney & Singapore & Temple & Tokyo & Taipei & Cairo & Brooklyn & Maiden's & \multirow{2}{1cm}{Average mAP}\\
& bridge & arch & familia & bridge & city & heaven & skytree & 101 & tower & bridge  & tower\\
\hline
\hline
\textsf{K-NN} \cite{im2gps} & 46.75 & 32.03 & 33.55 & 38.93 & 29.83 & 54.33 & 17.21 & 32.39 & 44.37 & 22.56 & 64.58 & 38.78\\
\hline
\textsf{LF+SVM} \cite{Zhu-MM08} & 40.52 & 52.35 & 45.78 & 36.09 & 29.17 & 61.23 & 38.96 & 15.99 & 34.55 & 33.58 & 62.33 & 40.96\\
\hline
\textsf{DRLR} \cite{Deorsch:paris} & 52.45 & 35.65 & 51.48 & 34.58 & 40.33 & 62.45 & 42.33 & 40.24 & 44.58 & 38.34 & 57.35 & 45.52\\
\hline
\textsf{GIANT} \cite{Fang-MM13} & 53.15 & 42.44 & 50.89 & 39.15 & 42.36 & 64.37 & 46.73 & 43.59 & 47.21 & 37.82 & 64.58 & 49.29\\
\hline
\textsf{PAMQE} \cite{YXLZ-MM15} & 57.83 & 64.76 & 57.33 & 45.59 & 63.48 & 68.38 & 60.48 & 59.32 & 70.04 & 54.58 & 72.38 & 61.29\\
\hline
\textsf{MMHG} \cite{Landmark-hypergraph} & 49.44 & 41.91 & 47.15 &40.91 &41.02 &60.94 & 49.27 & 41.37 & 44.29 & 39.84 & 60.67 & 46.31\\
\hline
Ours & \textbf{60.25} & \textbf{67.42} & \textbf{59.64} & \textbf{49.22} & \textbf{65.77} & \textbf{70.14} & \textbf{62.25} & \textbf{61.25} & \textbf{71.25} & \textbf{58.77} & \textbf{74.05} & \textbf{63.64}\\
\hline
\end{tabular}\label{tab:mAP-flickr}
\end{table*}

\subsection{Comparison with Baselines}\label{ssec:compare_baseline}

This experiment is conducted to evaluate the effectiveness of our method in capturing the high-level discriminative representations for landmark retrieval by comparison with strong deep baselines. We use CNN pre-trained on ImageNet with the hope to reduce the chance of over-fitting to certain scenes. For the efficiency of the feature extraction process, we use Caffe library \cite{Caffe}. For the fine-tuning process, we take all images in the training set for each category as the input, and use the CNN to perform a prediction for each image. Please note that for the proposed C-CNN, training images are clustered into pseudo labels. The fully-connected layer generates a feature vector of 4096-dim which can be used to train one-vs-all SVMs for each scene category. In testing, an input image goes through the network model corresponding to each baseline, and its deep features are used to predict its scene classification for each linear SVM model, then we assign the one with highest confidence score.

In Table \ref{tab:classification}, we report classification accuracies of baselines on the testing set of Flickr and Picasa datasets. We can see that C-CNN model outperforms CNN-F and its variants, \ie CNN-F+Average and CNN-F+FV. This verifies that fine-tuning CNNs on pseudo labels with user information is helpful in constructing multi-query set from which more discriminative features can be generated. This is mainly because the latent factors revealed by user-landmark matrix factorization provide semantics to describe landmark photos from different perspectives. On the other hand, the proposed C-CNN with Fisher Vector pooling is superior to state-of-the-art deep baselines, Deep Places Features and MetaObject-CNN. The main reason is deep places features are trained from a very large scene-centric dataset which contains photos from search engines with wide-ranging diversity and density. However, these trainable parameters cannot be easily applied into social media data without fine-tuning. MetaObject-CNN uses a region proposal technique to generate a set of patches potentially containing objects, and apply a pre-trained CNN to extract generic deep features from these patches. Consequently, the MetaObject-CNN is inferior to our model not only because its lack of fine-tuning but also because the region proposal step has no guarantee to generate highly discriminative patches in describing landmarks. By contrast, the proposed C-CNN with Fisher Vector with a multi-query set as input can produce discriminative representations by encoding multiple deep features in their higher order.

An illustration of feature responses from C-CNN is shown in Fig. \ref{fig:feature_map}. We can see that C-CNN is able to use the most informative region as the highest response during prediction. For example, for London clock tower, highest responses come from tower pin and circular clock plate.

\begin{table*}\tiny
\centering
\caption{The classification accuracies/precision by different baselines against Flickr and Picasa Web Album datasets.}
\begin{tabular}{c|c|c|c|c|c|c}\hline
Landmark & CNN-F+Average & CNN-F+FV & C-CNN+Average & Deep Places  Features \cite{PlacesNIPS14} & MetaObject-CNN \cite{DeepCNN-Scene} & Ours\\
\hline\hline
Flickr & 54.32$\pm$0.14 & 58.42$\pm$0.76  & 58.61$\pm$0.78 & 60.68$\pm$0.92  &  59.29$\pm$0.33 &\textbf{64.18$\pm$0.88}\\
\hline
Picasa & 56.79$\pm$0.37  & 57.22$\pm$0.92 &56.59$\pm$0.85 & 57.79$\pm$0.99 & 57.18$\pm$0.31  &\textbf{61.23$\pm$0.76}\\
\hline
\end{tabular}\label{tab:classification}
\end{table*}

\subsection{Cross Dataset Generalization}

\begin{figure}[hbt]
\centering
\begin{tabular}{cc}
\includegraphics[width=4cm]{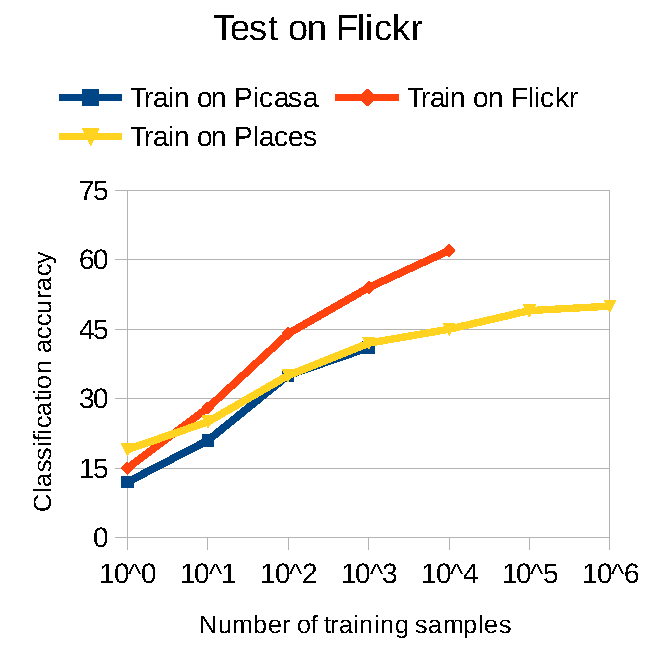}&
\includegraphics[width=4cm]{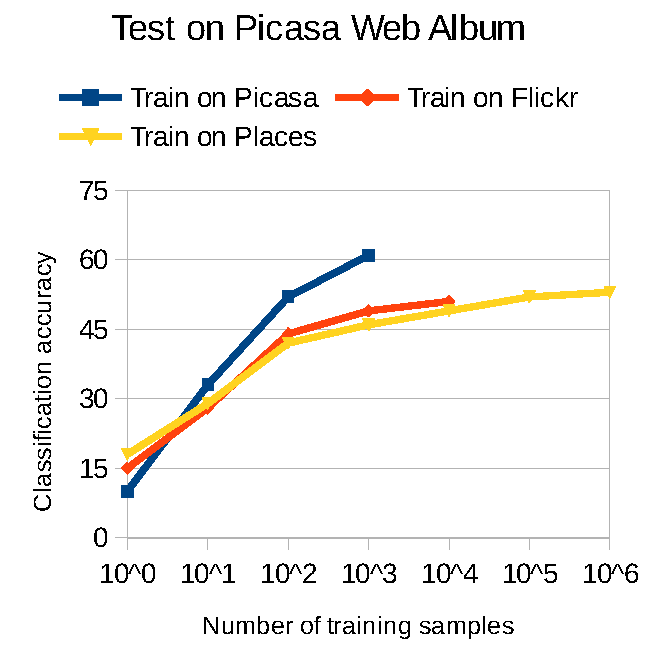}\\
\end{tabular}
\caption{Cross dataset generalization of training and testing on different datasets. See text for details. }\label{fig:cross_dataset}
\end{figure}

A well-known problem in deep feature learning for visual recognition is cross dataset generalization. It implies that training and testing across datasets generally results in a drop of performance due the dataset bias problem. In this case, the bias between datasets is due to the data distribution, density and diversity between the four datasets: ImageNet, Places dataset, Flickr, and Picasa Web Album. Specifically, for training on Flickr and Picasa, we use trainable parameters of CNN-F pre-trained on ImageNet and then fine-tuning parameters on target dataset. For training on Places dataset, we directly use the learned parameters since Places dataset is scene-centric. Fig. \ref{fig:cross_dataset} shows the classification results obtained from training and testing on different permutations on the four datasets. We can see that features extracted from a pre-trained ImageNet CNN-F fine-tuned on the training and testing from the same dataset provides the best performance for a fixed number of training examples. Although Places dataset is very large, its performance is inferior to CNN-F pre-trained on ImageNet with fine-tuning. A possible reason is features trained from ImageNet are generic and transferable to a target dataset.

\subsection{CNN Feature Training}

Considering the smaller size of our training dataset when compared with ILSVRC-2012, we controlled the over-fitting by using lower initial learning rates for the fine-tuned hidden layers (FC1 and FC2). The learning rate schedule for the last layer and hidden layers was: ($10^{-1}, 10^{-4}$) $\rightarrow$  ($10^{-3}, 10^{-4}$) $\rightarrow$ ($10^{-4}, 10^{-4}$) $\rightarrow$($10^{-5}, 10^{-5}$). Our network has the same dimensionality of the last hidden layer (FC2): 4096. This design choice is in accordance with state-of-the-art architecture \cite{NIPS12,Devil}. We further train three modifications of the CNN-F network, with lower dimensional FC2 layers of 2048, 1024, and 128 dimensions, respectively. To speed up training, all layers aside from FC2/FC3 were set to the those of the CNN-F net and a lower initial learning rate of $10^{-3}$ was used. The initial learning rate of FC2/FC3 was set to $10^{-2}$. Experimental results are given in Table \ref{tab:num_dim}. It can be seen that mAP values of two datasets decrease as the dimensionality of the last hidden layer becomes lower. One possible reason is 4096 is already rather compact, and further reduction on dimensionality would lead to under-performed results.

\begin{table}
\centering
\caption{mAP of two datasets with respect to modified architecture of CNN-F feature learning.}
\begin{tabular}{c|c|c}\hline
& Flickr (mAP) & Picasa (mAP)\\
\hline\hline
CNN-F (4096) & 63.64&63.62\\
\hline
CNN-F (2048) & 63.07 & 63.14\\
\hline
CNN-F (1024)& 62.45 & 61.95\\
\hline
CNN-F(128) & 61.21 & 60.82\\
\hline
\end{tabular}\label{tab:num_dim}
\end{table}

\subsection{Comparison with State-of-the-art Approaches}

\begin{figure}
\centering
\begin{tabular}{cc}
\includegraphics[width=3.5cm]{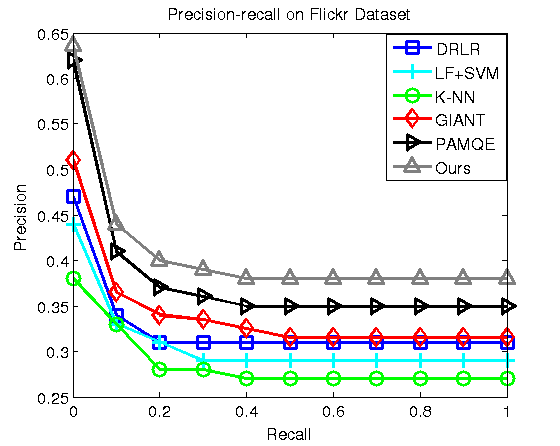}&
\includegraphics[width=3.5cm]{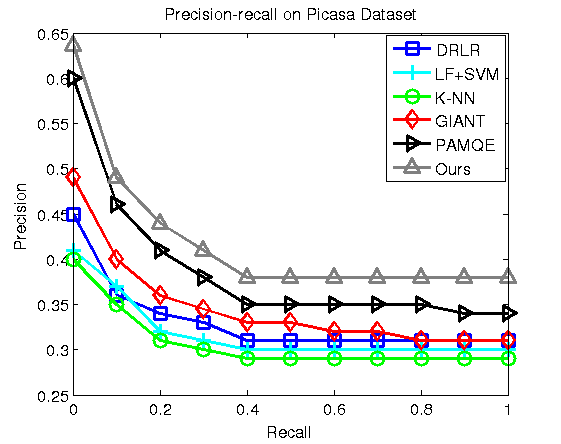}\\
\end{tabular}
\caption{Precision-recall results over Flickr and Picasa.}\label{fig:cision_recall}
\end{figure}

We evaluate mAP for our approach and competitors. For each landmark in Flickr and Picasa Web Album, we randomly sample 20 query photos for each landmark and perform landmark retrieval to return the ranking list, where we obtain the precision-recall value and plot a precision-recall curve, plotting precision $p(r)$ versus the recall $r$ in Fig. ~\ref{fig:cision_recall}, where our method outperforms others especially over our recent \textsf{PAMQE} \cite{YXLZ-MM15} to demonstrate the strength of high-level deep features learned over user-landmark latent subspace.
For each dataset, we also use mAP as the evaluation metric to examine the performance of each approach. In the case of duplicate landmark photos, the evaluation protocol of mean average precision (mAP) considers that a duplicate photo to the query is viewed as `'junk'' \cite{exp1}. This protocol is a common setting in object retrieval \cite{exp1,exp2}, and won't affect the retrieval accuracy values.

For a query $q$, the average precision (AP) is defined as $AP(q)=\frac{1}{L_q}\sum_{r=1}^n P_q(r) \theta_q(r)$, where $L_q$ is the ground-truth neighbors of query $q$ in database, $n$ is the number of photos in ranked list, we set $n = 100$, $P_q(r)$ denotes the precision of the top $r$ retrieved photos, and $\theta_q(r)=1$ if the $r$-th retrieved photo is a ground-truth within this landmark category and $\theta_q(r)=0$, otherwise. In our case, given 20 query photos for each landmark, the mAP is defined over all 20 queries for each landmark as: $mAP=\frac{1}{20}\sum_{i=1}^{20} AP(q_i)$.
In the step of multi-query expansions, by default, the multi-query set $Q$ are composed of 40 recommended photos.

The compared landmark retrieval results with mAP values are shown in Table \ref{tab:mAP-flickr} and Table \ref{tab:mAP-picasa}. We observe that: (1) Our method and \textsf{PAMQE} outperform all competitors, due to the effectiveness of exploiting the complementary information of multiple queries, while our method is better than \textsf{PAMQE} due to the superiority of high-level features over mid-level pattern representation; (2) The large intra-class variance limits the performance of \textsf{LF+SVM} and \textsf{K-NN}, especially for the Flickr dataset; (3) \textsf{DRLR} detects discriminative regions from a single query photo, which degrades its performance when a query photo was shot from a bad viewpoint; and (4) \textsf{MMHG} works by developing multiple hypergraphs to identify the associations between landmark photos. However, they are still using low-level features such as color moments and local binary patterns, making them underperformed. Although \textsf{GIANT} also performs better, it only utilizes user-generated content to obtain visual attributes. This also demonstrates the need of exploiting user information to assist query understanding.

\begin{table*}\tiny
\centering
\caption{The mAP values ($\%$) of selected landmarks by different approaches against Picasa Web Album dataset.}
\begin{tabular}{c|c|c|c|c|c|c}\hline
Landmark & \textsf{K-NN} \cite{im2gps} & \textsf{LF+SVM} \cite{Zhu-MM08} & \textsf{DRLR} \cite{Deorsch:paris} & \textsf{GIANT} \cite{Fang-MM13}  & \textsf{PAMQE} \cite{YXLZ-MM15} & Ours\\
\hline\hline
London clock & 58.37 & 50.03 & 60.65 & 62.20 & 71.89 &\textbf{74.05}\\
\hline
Eiffel tower &  51.24 & 51.05 & 46.77 & 52.13 & 61.71 &\textbf{65.43}\\
\hline
Forbidden city & 44.04 & 55.28 & 58.57 & 58.94 & 70.68 &\textbf{72.74}\\
\hline
Opera house & 49.26 & 33.48 & 46.88 &  48.48 & 61.74 &\textbf{65.87}\\
\hline
Catalunya plaza & 24.87 & 12.37 & 21.67 & 23.23 & 42.83 &\textbf{48.24}\\
\hline
Chicago plaza & 35.48 & 38.38 & 33.76 & 38.74 & 50.84 &\textbf{55.39}\\
\hline
Average mAP& 43.87 & 40.09 & 44.73 & 47.28 & 59.87 & \textbf{63.62}\\
\hline
\end{tabular}\label{tab:mAP-picasa}
\end{table*}

\subsection{Comparison with Query Expansion Approaches}

\begin{figure}[hbt]
\centering
\begin{tabular}{c}
\includegraphics[width=7cm]{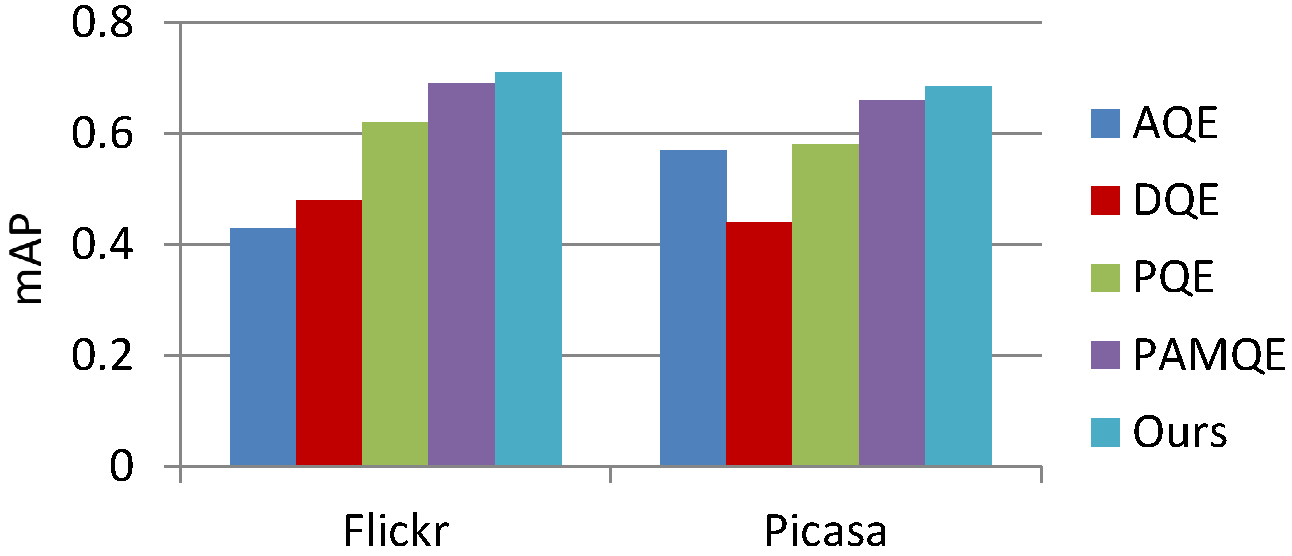}\\
\end{tabular}
\caption{Comparison of query expansion methods over mAPs. }\label{fig:QE-comp}
\end{figure}

Unlike \textsf{AQE} and \textsf{DQE} where plausible queries are selected via a low-level feature matching procedure, the proposed multi-query selection can complement a q-user by exploiting latent topics and information from users. While \textsf{PQE} uses similar scheme to expand a single query into a query set, its multi-query set is determined by manual selection.
To demonstrate the superiority of our method over existing multiple query methods that can be adopted for landmark retrieval, we perform comparison against several query expansion approaches: \textsf{AQE}, \textsf{DQE}, \textsf{PQE} and \textsf{PAMQE}. Results are shown in Fig.\ref{fig:QE-comp}, where we conclude that \textsf{PAMQE} outperforms \textsf{AQE}, \textsf{DQE}, and \textsf{PQE} by a large margin in terms of mAP values in two databases. However, it performs inferior to the proposed C-CNN with Fisher vector encoding, which can produce highly discriminative features for landmark retrieval.

\section{Conclusions}\label{sec:con}
In this paper, we propose a novel collaborative deep networks for robust landmark retrieval, which works over landmark latent factors to further generate the high-level semantic feature for both multi-query set and other landmark photos. Compared with both low-level feature and mid-level pattern representation based methods, our proposed method achieved state-of-the-art performance, validated by experimental results on real-world social landmark photo datasets associated with the user information.

{
\balance \vfill\eject
\bibliographystyle{named} \vfill\eject
\bibliography{ijcai15}
}

\end{document}